\documentclass[conference,a4paper]{IEEEtran}
\IEEEoverridecommandlockouts

\usepackage[hidelinks]{hyperref}
\usepackage[cmex10]{amsmath}%American Math Society(AMS) math formatting
\usepackage{amssymb,amsfonts}%AMS extra symbols and fonts
\usepackage{dblfloatfix}%fix double column figure ordering and placement

\usepackage[ruled,vlined]{algorithm2e}
\usepackage{graphicx}
\graphicspath{{Figures/PDF/}{Figures/PNG/}}

\usepackage{booktabs}
\usepackage{siunitx}
\usepackage[numbers,compress]{natbib}
\usepackage{texnames}
\usepackage{bm,bbm}
\usepackage{orcidlink}

\begin{document}

\title{\uppercase{Non-Registration Change Detection: A Novel Change Detection Task and Benchmark Dataset}
\thanks{All authors are affiliated with Hainan University, 570228 Haikou, China.}
}

\author{	\IEEEauthorblockN{Zhe Shan}
	% \IEEEauthorblockA{\textit{Hainan University}}\\
		%570228 Haikou, China\\
		%shanzard@163.com}
	\and
	\IEEEauthorblockN{Lei Zhou\orcidlink{0000-0003-0860-0724} \thanks{\IEEEauthorrefmark{1} Corresponding authors (Lei Zhou: leizhou@hainanu.edu.cn; Xia Xie: shelicy@hainanu.edu.cn).}\IEEEauthorrefmark{1}}
	% \IEEEauthorblockA{\textit{Hainan University}}\\
		%570228 Haikou, China\\
		%leizhou@hainanu.edu.cn}
	\and
	\IEEEauthorblockN{Liu Mao}
	% \IEEEauthorblockA{\textit{Hainan University}}\\
		%570228 Haikou, China\\
		%hnmaoliu@126.com}
    \and
	\IEEEauthorblockN{Shaofan Chen}
	% \IEEEauthorblockA{\textit{Hainan University}}\\
		%570228 Haikou, China\\
		%87143728@qq.com}
    \and
	\IEEEauthorblockN{Chuanqiu Ren}
	% \IEEEauthorblockA{\textit{Hainan University}}\\
		%570228 Haikou, China\\
		%terry@hainanu.edu.cn}
    \and
	\IEEEauthorblockN{Xia Xie \IEEEauthorrefmark{1}}
	% \IEEEauthorblockA{\textit{Hainan University}}\\
		%570228 Haikou, China\\
		%shelicy@hainanu.edu.cn}
}

\maketitle
\begin{abstract}
In this study, we propose a novel remote sensing change detection task, non-registration change detection, to address the increasing number of emergencies such as natural disasters, anthropogenic accidents, and military strikes. First, in light of the limited discourse on the issue of non-registration change detection, we systematically propose eight scenarios that could arise in the real world and potentially contribute to the occurrence of non-registration problems. Second, we develop distinct image transformation schemes tailored to various scenarios to convert the available registration change detection dataset into a non-registration version. Finally, we demonstrate that non-registration change detection can cause catastrophic damage to the state-of-the-art methods. Our code and dataset are available at https://github.com/ShanZard/NRCD.
\end{abstract}


\section{Introduction}

Remote sensing change detection (RSCD) identifies areas of semantic change by comparing images taken at different phases of the same location. With the rapid development of earth observation and computer vision technologies, RSCD has significantly contributed to disaster assessment~\citep{ZHENG2021112636}, urban management~\citep{XIAN20101676}, and resource monitoring~\citep{KENNEDY20091382}. The capture of massive amounts of various types of remote sensing data has promoted the development of data-driven algorithms~\citep{8278197,9146211,shan2025rossamh}.

In recent years, thanks to deep learning techniques~\citep{lecun2015deep,7486259}, CNN~\citep{9451544} or Transformer~\citep{9716741} based remote sensing change detection models have shown superior performance and gradually replaced traditional methods. Research scholars have proposed a large number of classical change detection datasets~\citep{Chen2020,shi21deeply,9780164,10145434,ZHANG20231,10744421} and methods~\citep{chen2021a,9632564,chen2022rdpnet,li2022cd,9883686,YeCD,Li_2023_A2Net,10034787}.~\citet{chen2021a} believe that objects with the same semantic concepts may display different spectral features at various temporal and spatial locations, and they propose a Bitemporal Image Transformer (BIT) to effectively model contexts within the spatiotemporal domain.~\citet{chen2022rdpnet} consider the deployment of algorithms on edge devices and propose the lightweight RDP-Net to achieve inference with fewer floating-point operations while enhancing attention to details such as bounding and small regions.~\citet{10443350} introduce the vision foundation model for change detection and demonstrate efficient sample learning comparable to that of semi-supervised change detection methods.~\citet{CHENG20241} propose a Content Cleansing Network (CCNet) that utilizes decoupled representation learning to separate an image's content and style features, effectively mitigating the effect of pseudo-change on accuracy. Due to the above state-of-the-art work efforts, change detection has significantly improved in both accuracy and speed in recent years.

\begin{figure}[t]
	\centering
	\includegraphics[width=\linewidth]{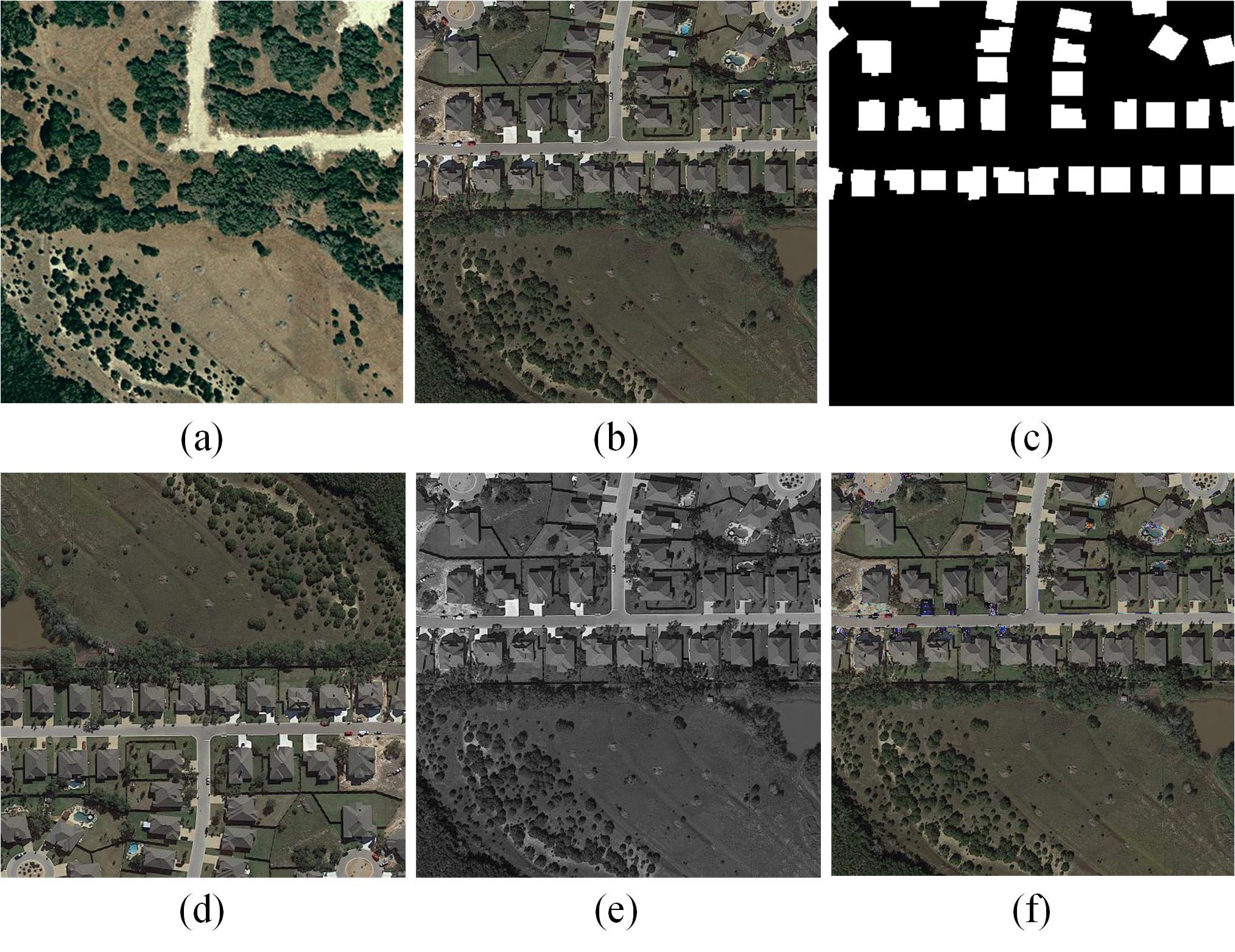}
	\caption{Here are examples of several types of non-registration change detection images: (a) The image captured at time $T_1$, representing the state before changes occur. (b) The image captured at time $T_2$, showing the state after changes have occurred. (c) The change area mask. (d) The flight paths at times $T_1$ and $T_2$ do not align. (e) Inconsistencies in image modality between $T_1$ and $T_2$. (f) The image at $T_2$ is affected by electromagnetic interference, and some areas are distorted.
    }\label{fig:fig1}
    \vspace{-15pt}
\end{figure}

To identify the object of change within a specific region, nearly all current change detection datasets and algorithms are based on co-registered remote sensing images taken from two different periods. In many real-world scenes, we often have access only to two non-registered images. For example, an image captured by a UAV following an emergency, such as a military strike or sudden natural disaster, can be challenging to align with a prior image taken by a remote sensing satellite. Fig.~\ref{fig:fig1} illustrates some non-registration change detection images. In such emergencies, we urgently need a rapid change detection algorithm to identify changes in critical ground objects, including land, mountains, water bodies, buildings, roads, etc. Within the scope of our survey, there is no current research that systematically discusses the issue of non-registration in remote sensing change detection and the significant damage it inflicts on state-of-the-art change detection methods. Only~\citet{JING202564} are concerned about the possible unalignment of viewpoints in different time-phase images, proposing ChangeRD and designing the adaptive perspective transformation (APT) module. This module can spatially transform intermediate layer features within the twin network, thereby mitigating the effect of unalignment on the change detection task. However, this work simulates only one non-registration state by randomly offsetting the vertices of registered bi-temporal images and computing the perspective transformation matrix to warp the entire image. Based on these reasons, we believe it is time to propose a new task, i.e., non-registration change detection, and explore the usability of various types of algorithms in emergencies.

In this study, we introduce a novel task: non-registration change detection (NRCD), based on real-world application requirements. First, we systematically examine eight non-registration patterns based on potential emergencies, including coordinate changes, color alterations, and encountering interference.  Second, we simulate these eight non-registration patterns by matrix computation and implement the conversion of an existing aligned dataset to a non-registration dataset. Finally, we demonstrate the disruption caused by this task to current algorithms and discuss the challenges of implementing non-registration change detection.

\section{Methodology}

\subsection{Preliminary}
RSCD is a basic and important task in remote sensing. Given a training set represented as $
\mathcal{D}_{\text{train}} = \left\{ \left( X_{i}^{T_{1}}, X_{i}^{T_{2}}, Y_{i} \right) \right\}_{i=1}^{N_{\text{train}}}$, where$ X_{i}^{T_{1}}, X_{i}^{T_{2}} \in \mathcal{R}^{H \times W \times C}$ is the $i$-th multi-temporal image pair acquired in \( T_{1} \) and \( T_{2} \), respectively, and \( Y_{i} \in \{0,1,\cdots,C\}^{H \times W} \) is the corresponding label, the goal of RSCD is to train a change detector \( \mathcal{F}_{\theta}\) on \( \mathcal{D}_{\text{train}} \) that can predict change maps reflecting accurate change information on new sets \( \mathcal{D}_{\text{test}} \). In previous RSCD settings, images $X_{i}^{T_{1}}, X_{i}^{T_{2}}$ in  $\mathcal{D}_{\text{train}}$ and  \( \mathcal{D}_{\text{test}} \) are registered.

In NRCD, the challenge of constructing change detector \( \mathcal{F}_{\theta}\) is enormous, as the image in \( \mathcal{D}_{\text{test}} \)  is unable to complete the registration. Currently, all existing RSCD datasets are registered. To explore non-registration change detection (NRCD), a straightforward and efficient method for converting a registration dataset to a non-registration dataset is by simulating and synthesizing. To simulate the non-registration scenario, we assume that the $T_1$ image serves as a perfect reference, and all transformations are applied exclusively to the $T_2$ image. The labels remain unchanged. In this context, $T_1$ denotes a ground image that was precisely scanned prior to an emergency event, while $T_2$ refers to a ground image captured promptly following the event. The labels represent the changes of the ground objects in $T_1$ after the emergency event.

\subsection{Non-Registration Change Detection}
\begin{figure*}[htpb]
	\centering
	\includegraphics[width=0.9\linewidth]{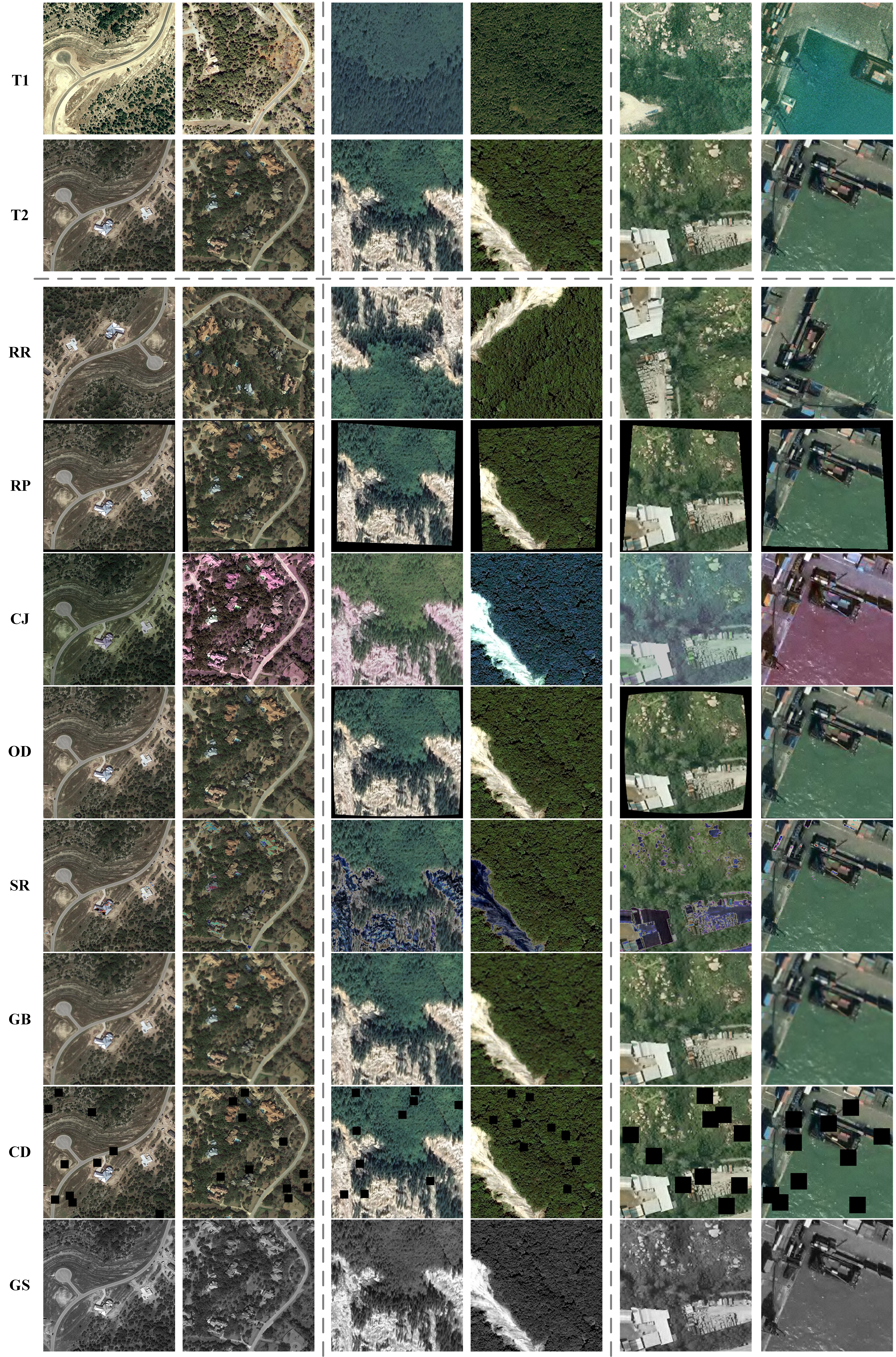}
	\caption{The visualization results of the proposed eight non-registration scenarios are shown in the LEVIR-CD, GVLM, and SYSU-CD datasets (from left to right). The abbreviations RR, RP, CJ, OD, SR, GB, CD, and GS stand for random rotation, random perspective, color jitter, optical distortion, solarization, Gaussian blur, coarse dropout, and grayscaling, respectively.}\label{fig:Densities}
\end{figure*}
Fig.~\ref{fig:Densities} illustrates the eight non-registration scenes proposed in this paper with the LEVIR-CD~\citep{Chen2020}, GVLM~\citep{ZHANG20231}, and SYSU-CD~\citep{shi21deeply} datasets in their non-registration versions. These non-registration images are generated by applying image transformations to the original dataset. These transformations are efficiently executed using the Python libraries torchvision~\citep{paszke2019pytorch} and albumentations~\citep{info11020125}, so further algorithm description is unnecessary. We focus on the motivation behind each transformation, and its real-world application is detailed in the following.

\subsubsection{Random Rotation}

Remote sensing images differ from natural images in that they are not influenced by gravity, making it impossible to observe orientation from the images themselves directly. In emergencies, an image captured at time $T_2$ may lack orientation coordinates and be misaligned with an image taken at time $T_1$. To simulate this scenario of non-registration change detection, we can rotate the $T_2$ image.

\subsubsection{Random Perspective}

In emergencies, an airplane may not be able to reach the same position to capture images and might have a certain inclination when photographing ground objects. This inclination causes the image to exhibit distortion, where nearby objects appear larger and distant objects appear smaller, complicating detection. We can simulate this non-registration effect through perspective transformation.

\subsubsection{Color Jitter}

Due to various factors such as the camera sensor, light source color temperature, and post-processing algorithms, the color of ground photos taken at different stages may vary significantly. This type of non-registration can be simulated by randomly jittering the brightness, contrast, saturation, and hue of the image.

\subsubsection{Optical Distortion}
In some emergencies, a fisheye camera may be selected for its wider field of view, but it inherently produces optical distortion. To simulate this non-registration situation, we can apply a similar optical distortion to the $T_2$ image.

\subsubsection{Solarization}
During ground observation, remote sensing cameras are susceptible to interference from highly reflective ground objects or counter-surveillance equipment, resulting in pixel distortion in certain areas. To simulate this situation, we can apply solarization transformation to the $T_2$ image. Solarization is a phenomenon in photography where the image recorded on a negative or photographic print is wholly or partially reversed in tone.

\subsubsection{Gaussian Blur}

In emergency situations, there may be blurring of the image caused by various factors such as inability to focus, equipment shake, and atmospheric disturbances.  We can simulate this effect using Gaussian Blur by setting the Gaussian kernel size and $\sigma$ value to achieve different degrees of blur.

\begin{table*}[hbt]
    \centering
 \fontsize{7pt}{8pt}\selectfont
    \caption{Quantitative evaluation results of different algorithms and non-registration scenarios on the LEVIR-CD dataset.}\label{tab:MyTable}
    \begin{tabular}{lccccccccc}
        \toprule
        
        \textbf{Type}  & \textbf{None} & \textbf{RR}& \textbf{RP}& \textbf{CJ}& \textbf{OD}& \textbf{SR}& \textbf{GB}& \textbf{CD}& \textbf{GS}\\         
        \midrule
    BIT~\citep{chen2021a}          & 78.76 & 7.23 (-71.53) & 39.26 (-39.50) & 64.66 (-14.10) & 58.75 (-20.01) & 70.14 (-1.62) & 58.78 (-19.98) & 72.62 (-6.06) & 62.63 (-16.13) \\
    DMINet~\citep{10034787}        & 74.36 & 5.49 (-68.87) & 37.60 (-36.76) & 38.79 (-35.57) & 55.43 (-18.93) & 63.41 (-10.95) & 64.63 (-10.27) & 70.14 (-5.78) & 60.04 (-14.68) \\
    ChangerFormer~\citep{9883686}  & 76.16 & 6.49 (-69.67) & 32.21 (-43.95) & 27.66 (-48.50) & 56.02 (-20.14) & 65.33 (-10.83) & 74.97 (-1.19) & 69.73 (-6.43) & 48.14 (-28.02) \\
    TFI-GR~\citep{li2022cd}        & 81.41 & 7.20 (-74.21) & 38.49 (-42.92) & 66.86 (-14.55) & 59.57 (-21.84) & 72.72 (-8.69) & 76.48 (-4.93) & 76.32 (-5.09) & 66.66 (-14.75) \\
    RDPNet~\citep{chen2022rdpnet}  & 68.13 & 6.09 (-62.04) & 35.35 (-32.78) & 38.59 (-29.54) & 52.43 (-14.30) & 56.40 (-10.73) & 67.46 (-1.68) & 62.18 (-4.05) & 50.72 (-17.41) \\
    SNUNet~\citep{9355573}         & 78.49 & 5.89 (-72.60) & 39.26 (-39.23) & 42.40 (-36.09) & 58.02 (-20.47) & 68.63 (-10.06) & 67.00 (-11.49) & 73.42 (-5.07) & 63.25 (-15.24) \\
    AFCF3D~\citep{YeCD}            & 79.01 & 7.59 (-71.42) & 42.64 (-36.37) & 53.41 (-25.60) & 60.52 (-18.51) & 70.80 (-1.79) & 67.27 (-1.74) & 75.38 (-6.37) & 60.15 (-18.86) \\
    CGNet~\citep{10234560}         & 83.34 & 8.12 (-75.22) & 36.93 (-46.41) & 73.84 (-9.50) & 60.81 (-22.53) & 75.67 (-12.33) & 76.99 (-9.65) & 77.34 (-10.00) & 72.55 (-10.21) \\
    HCGMNet~\citep{10283341}       & 84.01 & 7.71 (-76.30) & 40.36 (-43.65) & 74.85 (-9.16) & 61.22 (-22.79) & 75.96 (-11.95) & 76.29 (-7.72) & 77.84 (-11.18) & 72.82 (-11.19) \\
    \textbf{Mean}                  & 78.19 & 6.87 (-71.31) & 38.01 (-40.17) & 53.45 (-24.73)& 58.09 (-19.95)  & 68.78 (-8.77) & 69.99 (-7.63) & 72.77 (-6.67) & 61.88 (-16.28) \\
        \bottomrule                
    \end{tabular}  
    \vspace{-15pt}
\end{table*}
\subsubsection{Coarse Dropout}
During remote sensing imaging, various disturbances can easily result in incomplete data, leading to missing information in certain areas. This issue is particularly problematic in emergency situations, where it is not feasible to rescan the ground to obtain complete data. We can simulate this non-registration situation using coarse dropout, which allows for the simulation of various incomplete images by adjusting the size, number, and fill value of the dropout holes.

\subsubsection{Grayscale}
Some remote sensing imaging techniques only produce single-channel grayscale maps, such as PAN and SAR images. Although these images lack color features, they still capture certain object features and can be used for before-and-after change comparisons. We can simulate this non-registration situation to some extent by converting the $T_2$ images to grayscale.

\section{Experiment and Discussion}

\subsection{Evaluation Metrics}\label{3.2}

In this study, we use  IoU to evaluate the performance of the various algorithms. IoU represents the ratio between the intersection and the union of prediction results and labels.
% Following are the calculation formulae for evaluation metrics:

% \begin{equation}
% IoU=\frac{TP}{TP+FP+FN}
% \end{equation}

% $TP$, $FP$, and $FN$ are the numbers of true positives, false positives, and false negatives, respectively. 

\subsection{Experimental Settings}\label{3.3}
All the experiments in this study are performed on NVIDIA GeForce RTX 4090 (24GB memory), and the Pytorch deep learning library are used to construct, train, and test the model. We uniformly adopt cross-entropy loss as the objective function.  The optimizer used in the study is AdamW, and the batch size is set to 8. The number of epochs is set to 200. The initial learning rate is $5\times10^{-4}$. Additionally, we do not use any data augmentation in the training process to fairly verify the destructiveness of various types of non-registration to the algorithm.

\subsection{Analysis of Experiment Results}
The quantitative evaluation results of each method on the LEVIR-CD dataset are presented in Table~\ref{tab:MyTable}. None represents the original registration dataset, with the differences for each case from None indicated in parentheses. The value in parentheses represents how much it has decreased compared to None. The quantitative results indicate that the eight non-registration scenarios proposed in this study significantly affect all algorithms.

First, the most significant impact on algorithms arises from random rotations. This effect is easily explained, as most algorithms assess changes by comparing feature maps, and the feature map of $T_2$ after rotation cannot be directly compared with that of $T_1$. Second, changes in coordinates have a greater detrimental effect than changes in color space. Specifically, alterations to an object's position are more disruptive to the algorithm than changes in feature information. For instance, RR and RP have a more substantial impact on algorithmic accuracy than CJ. Additionally, local feature distortions, incompleteness, and blurring, which result in information loss, have a relatively minor effect on algorithms. Lastly, the robustness of different algorithms varies across scenarios, with no clear pattern to determine which method is best suited for a particular scenario. Finally, 

\subsection{Discussion }

The above experiments have highlighted the significant challenges that non-registration scenes present to existing change detection algorithms. However, these data are merely simulated syntheses and still fail to adequately represent the challenges of real non-registration scenarios. For instance, if the image in $T_2$ is a real SAR or PAN image, its impact on model accuracy is significantly greater than that represented by a single-channel grayscale image, which in this case is true cross-modal. Additionally, various real-world physical effects, such as atmospheric conditions and radiation, are difficult to replicate through simulation synthesis. Finally, although the synthesized scenes may be relatively simple, they have actually posed significant challenges to current algorithms, so we believe that using synthesis for discussion is also of great significance.

\section{Conclusions}

In this work, we emphasize the challenge posed by non-registration scenes to existing change detection algorithms. Based on potential non-registration situations arising from emergency events, we propose eight types of non-registration transformations to transform existing datasets into non-registration datasets. Further, we conduct the SOTA existing change detection algorithms on the synthesized non-registration detection dataset, and the experimental results show that these scenarios cause catastrophic damage to the existing algorithms. We believe this is preliminary but fundamental work that potentially serves the growing number of emergencies.

\noindent\textbf{Acknowledgement.} This work was supported by the National Natural Science Foundation of China (No. 62362023, No. 62402354), and the Key Research and Development Project of Hainan Province (ZDYF2024GXJS313, ZDYF2024GXJS262).

\small
\bibliographystyle{IEEEtranN}
\bibliography{references}

% Generated by IEEEtranN.bst, version: 1.14 (2015/08/26)
\begin{thebibliography}{32}
\providecommand{\natexlab}[1]{#1}
\providecommand{\url}[1]{#1}
\csname url@samestyle\endcsname
\providecommand{\newblock}{\relax}
\providecommand{\bibinfo}[2]{#2}
\providecommand{\BIBentrySTDinterwordspacing}{\spaceskip=0pt\relax}
\providecommand{\BIBentryALTinterwordstretchfactor}{4}
\providecommand{\BIBentryALTinterwordspacing}{\spaceskip=\fontdimen2\font plus
\BIBentryALTinterwordstretchfactor\fontdimen3\font minus \fontdimen4\font\relax}
\providecommand{\BIBforeignlanguage}[2]{{%
\expandafter\ifx\csname l@#1\endcsname\relax
\typeout{** WARNING: IEEEtranN.bst: No hyphenation pattern has been}%
\typeout{** loaded for the language `#1'. Using the pattern for}%
\typeout{** the default language instead.}%
\else
\language=\csname l@#1\endcsname
\fi
#2}}
\providecommand{\BIBdecl}{\relax}
\BIBdecl

\bibitem[Zheng et~al.(2021)Zheng, Zhong, Wang, Ma, and Zhang]{ZHENG2021112636}
Z.~Zheng, Y.~Zhong, J.~Wang, A.~Ma, and L.~Zhang, ``Building damage assessment for rapid disaster response with a deep object-based semantic change detection framework: From natural disasters to man-made disasters,'' \emph{Remote Sensing of Environment}, vol. 265, p. 112636, 2021.

\bibitem[Xian and Homer(2010)]{XIAN20101676}
G.~Xian and C.~Homer, ``Updating the 2001 national land cover database impervious surface products to 2006 using landsat imagery change detection methods,'' \emph{Remote Sensing of Environment}, vol. 114, no.~8, pp. 1676--1686, 2010.

\bibitem[Kennedy et~al.(2009)Kennedy, Townsend, Gross, Cohen, Bolstad, Wang, and Adams]{KENNEDY20091382}
R.~E. Kennedy, P.~A. Townsend, J.~E. Gross, W.~B. Cohen, P.~Bolstad, Y.~Wang, and P.~Adams, ``Remote sensing change detection tools for natural resource managers: Understanding concepts and tradeoffs in the design of landscape monitoring projects,'' \emph{Remote Sensing of Environment}, vol. 113, no.~7, pp. 1382--1396, 2009.

\bibitem[Bai et~al.(2018)Bai, Xu, Zhou, Xing, Bai, and Zhou]{8278197}
X.~Bai, F.~Xu, L.~Zhou, Y.~Xing, L.~Bai, and J.~Zhou, ``Nonlocal similarity based nonnegative tucker decomposition for hyperspectral image denoising,'' \emph{IEEE Journal of Selected Topics in Applied Earth Observations and Remote Sensing}, vol.~11, no.~3, pp. 701--712, 2018.

\bibitem[Zhou et~al.(2020)Zhou, Zhang, Wang, Bai, Tong, Zhang, Zhou, and Hancock]{9146211}
L.~Zhou, X.~Zhang, J.~Wang, X.~Bai, L.~Tong, L.~Zhang, J.~Zhou, and E.~Hancock, ``Subspace structure regularized nonnegative matrix factorization for hyperspectral unmixing,'' \emph{IEEE Journal of Selected Topics in Applied Earth Observations and Remote Sensing}, vol.~13, pp. 4257--4270, 2020.

\bibitem[Shan et~al.(2025)Shan, Liu, Zhou, Yan, Wang, and Xie]{shan2025rossamh}
Z.~Shan, Y.~Liu, L.~Zhou, C.~Yan, H.~Wang, and X.~Xie, ``Ros-sam: High-quality interactive segmentation for remote sensing moving object,'' \emph{arXiv preprint arXiv:2503.12006}, 2025.

\bibitem[LeCun et~al.(2015)LeCun, Bengio, and Hinton]{lecun2015deep}
Y.~LeCun, Y.~Bengio, and G.~Hinton, ``Deep learning,'' \emph{Nature}, vol. 521, no. 7553, pp. 436--444, 2015.

\bibitem[Zhang et~al.(2016)Zhang, Zhang, and Du]{7486259}
L.~Zhang, L.~Zhang, and B.~Du, ``Deep learning for remote sensing data: A technical tutorial on the state of the art,'' \emph{IEEE Geoscience and Remote Sensing Magazine}, vol.~4, no.~2, pp. 22--40, 2016.

\bibitem[Li et~al.(2022{\natexlab{a}})Li, Liu, Yang, Peng, and Zhou]{9451544}
Z.~Li, F.~Liu, W.~Yang, S.~Peng, and J.~Zhou, ``A survey of convolutional neural networks: Analysis, applications, and prospects,'' \emph{IEEE Transactions on Neural Networks and Learning Systems}, vol.~33, no.~12, pp. 6999--7019, 2022.

\bibitem[Han et~al.(2023{\natexlab{a}})Han, Wang, Chen, Chen, Guo, Liu, Tang, Xiao, et~al.]{9716741}
K.~Han, Y.~Wang, H.~Chen, X.~Chen, J.~Guo, Z.~Liu, Y.~Tang, A.~Xiao \emph{et~al.}, ``A survey on vision transformer,'' \emph{IEEE Transactions on Pattern Analysis and Machine Intelligence}, vol.~45, pp. 87--110, 2023.

\bibitem[Chen and Shi(2020)]{Chen2020}
H.~Chen and Z.~Shi, ``A spatial-temporal attention-based method and a new dataset for remote sensing image change detection,'' \emph{Remote Sensing}, vol.~12, no.~10, 2020.

\bibitem[Shi et~al.(2021)Shi, Liu, Li, Liu, Wang, and Zhang]{shi21deeply}
Q.~Shi, M.~Liu, S.~Li, X.~Liu, F.~Wang, and L.~Zhang, ``A deeply supervised attention metric-based network and an open aerial image dataset for remote sensing change detection,'' \emph{IEEE Transactions on Geoscience and Remote Sensing}, pp. 1--16, 2021.

\bibitem[Liu et~al.(2022)Liu, Chai, Deng, and Liu]{9780164}
M.~Liu, Z.~Chai, H.~Deng, and R.~Liu, ``A cnn-transformer network with multiscale context aggregation for fine-grained cropland change detection,'' \emph{IEEE Journal of Selected Topics in Applied Earth Observations and Remote Sensing}, vol.~15, pp. 4297--4306, 2022.

\bibitem[Holail et~al.(2023)Holail, Saleh, Xiao, and Li]{10145434}
S.~Holail, T.~Saleh, X.~Xiao, and D.~Li, ``Afde-net: Building change detection using attention-based feature differential enhancement for satellite imagery,'' \emph{IEEE Geoscience and Remote Sensing Letters}, vol.~20, pp. 1--5, 2023.

\bibitem[Zhang et~al.(2023)Zhang, Yu, Pun, and Shi]{ZHANG20231}
X.~Zhang, W.~Yu, M.-O. Pun, and W.~Shi, ``Cross-domain landslide mapping from large-scale remote sensing images using prototype-guided domain-aware progressive representation learning,'' \emph{ISPRS Journal of Photogrammetry and Remote Sensing}, vol. 197, pp. 1--17, 2023.

\bibitem[Yu et~al.(2024)Yu, Zhang, Gloaguen, Xiang~Zhu, and Ghamisi]{10744421}
W.~Yu, X.~Zhang, R.~Gloaguen, X.~Xiang~Zhu, and P.~Ghamisi, ``Minenetcd: A benchmark for global mining change detection on remote sensing imagery,'' \emph{IEEE Transactions on Geoscience and Remote Sensing}, vol.~62, pp. 1--16, 2024.

\bibitem[Chen et~al.(2022{\natexlab{a}})Chen, Qi, and Shi]{chen2021a}
H.~Chen, Z.~Qi, and Z.~Shi, ``Remote sensing image change detection with transformers,'' \emph{IEEE Transactions on Geoscience and Remote Sensing}, vol.~60, pp. 1--14, 2022.

\bibitem[Guo et~al.(2022)Guo, Zhang, Zhu, Zhong, and Zhang]{9632564}
Q.~Guo, J.~Zhang, S.~Zhu, C.~Zhong, and Y.~Zhang, ``Deep multiscale siamese network with parallel convolutional structure and self-attention for change detection,'' \emph{IEEE Transactions on Geoscience and Remote Sensing}, vol.~60, pp. 1--12, 2022.

\bibitem[Chen et~al.(2022{\natexlab{b}})Chen, Pu, Yang, Tang, and Xu]{chen2022rdpnet}
H.~Chen, F.~Pu, R.~Yang, R.~Tang, and X.~Xu, ``Rdp-net: Region detail preserving network for change detection,'' \emph{IEEE Transactions on Geoscience and Remote Sensing}, vol.~60, pp. 1--10, 2022.

\bibitem[Li et~al.(2022{\natexlab{b}})Li, Tang, Wang, and Zomaya]{li2022cd}
Z.~Li, C.~Tang, L.~Wang, and A.~Y. Zomaya, ``Remote sensing change detection via temporal feature interaction and guided refinement,'' \emph{IEEE Transactions on Geoscience and Remote Sensing}, vol.~60, pp. 1--11, 2022.

\bibitem[Bandara and Patel(2022)]{9883686}
W.~G.~C. Bandara and V.~M. Patel, ``A transformer-based siamese network for change detection,'' in \emph{IEEE International Geoscience and Remote Sensing Symposium}, 2022, pp. 207--210.

\bibitem[Ye et~al.(2023)Ye, Wang, Zhou, Lei, Fan, and Qin]{YeCD}
Y.~Ye, M.~Wang, L.~Zhou, G.~Lei, J.~Fan, and Y.~Qin, ``Adjacent-level feature cross-fusion with 3-d cnn for remote sensing image change detection,'' \emph{IEEE Transactions on Geoscience and Remote Sensing}, vol.~61, pp. 1--14, 2023.

\bibitem[Li et~al.(2023)Li, Tang, Liu, Zhang, et~al.]{Li_2023_A2Net}
Z.~Li, C.~Tang, X.~Liu, W.~Zhang \emph{et~al.}, ``Lightweight remote sensing change detection with progressive feature aggregation and supervised attention,'' \emph{IEEE Transactions on Geoscience and Remote Sensing}, vol.~61, pp. 1--12, 2023.

\bibitem[Feng et~al.(2023)Feng, Jiang, Xu, and Zheng]{10034787}
Y.~Feng, J.~Jiang, H.~Xu, and J.~Zheng, ``Change detection on remote sensing images using dual-branch multilevel intertemporal network,'' \emph{IEEE Transactions on Geoscience and Remote Sensing}, vol.~61, pp. 1--15, 2023.

\bibitem[Ding et~al.(2024)Ding, Zhu, Peng, Tang, Yang, and Bruzzone]{10443350}
L.~Ding, K.~Zhu, D.~Peng, H.~Tang, K.~Yang, and L.~Bruzzone, ``Adapting segment anything model for change detection in vhr remote sensing images,'' \emph{IEEE Transactions on Geoscience and Remote Sensing}, vol.~62, pp. 1--11, 2024.

\bibitem[Cheng et~al.(2024)Cheng, He, Li, Yang, and Zhang]{CHENG20241}
M.~Cheng, W.~He, Z.~Li, G.~Yang, and H.~Zhang, ``Harmony in diversity: Content cleansing change detection framework for very-high-resolution remote-sensing images,'' \emph{ISPRS Journal of Photogrammetry and Remote Sensing}, vol. 218, pp. 1--19, 2024.

\bibitem[Jing et~al.(2025)Jing, Chi, Li, and Wang]{JING202564}
W.~Jing, K.~Chi, Q.~Li, and Q.~Wang, ``Changerd: A registration-integrated change detection framework for unaligned remote sensing images,'' \emph{ISPRS Journal of Photogrammetry and Remote Sensing}, vol. 220, pp. 64--74, 2025.

\bibitem[Paszke et~al.(2019)Paszke, Gross, Massa, Lerer, Bradbury, Chanan, Killeen, Lin, Gimelshein, Antiga, et~al.]{paszke2019pytorch}
A.~Paszke, S.~Gross, F.~Massa, A.~Lerer, J.~Bradbury, G.~Chanan, T.~Killeen, Z.~Lin, N.~Gimelshein, L.~Antiga \emph{et~al.}, ``Pytorch: An imperative style, high-performance deep learning library,'' \emph{Advances in Neural Information Processing Systems}, vol.~32, 2019.

\bibitem[Buslaev et~al.(2020)Buslaev, Iglovikov, Khvedchenya, Parinov, Druzhinin, and Kalinin]{info11020125}
A.~Buslaev, V.~I. Iglovikov, E.~Khvedchenya, A.~Parinov, M.~Druzhinin, and A.~A. Kalinin, ``Albumentations: Fast and flexible image augmentations,'' \emph{Information}, vol.~11, no.~2, 2020.

\bibitem[Fang et~al.(2022)Fang, Li, Shao, and Li]{9355573}
S.~Fang, K.~Li, J.~Shao, and Z.~Li, ``Snunet-cd: A densely connected siamese network for change detection of vhr images,'' \emph{IEEE Geoscience and Remote Sensing Letters}, vol.~19, pp. 1--5, 2022.

\bibitem[Han et~al.(2023{\natexlab{b}})Han, Wu, Guo, Hu, Li, and Chen]{10234560}
C.~Han, C.~Wu, H.~Guo, M.~Hu, J.~Li, and H.~Chen, ``Change guiding network: Incorporating change prior to guide change detection in remote sensing imagery,'' \emph{IEEE Journal of Selected Topics in Applied Earth Observations and Remote Sensing}, vol.~16, pp. 8395--8407, 2023.

\bibitem[Han et~al.(2023{\natexlab{c}})Han, Wu, and Du]{10283341}
C.~Han, C.~Wu, and B.~Du, ``Hcgmnet: A hierarchical change guiding map network for change detection,'' in \emph{IEEE International Geoscience and Remote Sensing Symposium}, 2023, pp. 5511--5514.

\end{thebibliography}

\end{document}